\pdfoutput=1

\documentclass[11pt]{article}

\usepackage[final]{acl}
\usepackage{tikz}
\usepackage{times}
\usepackage{latexsym}
\usepackage{subfigure}
\usepackage{graphicx}    
\usepackage{caption}     
\usepackage{subcaption}  
\usepackage[T1]{fontenc}

\usepackage[utf8]{inputenc}

\usepackage{microtype}

\usepackage{inconsolata}

\usepackage{graphicx}
\usepackage{todonotes}
\usepackage{color}
\definecolor{purple}{rgb}{1,0,1}

\usepackage{pifont}
\newcommand{\cmark}{\ding{51}}%
\newcommand{\xmark}{\ding{55}}%

\newcount\Comments  
\newcommand{\kibitz}[2]{\ifnum\Comments=1{\color{#1}{#2}}\fi}

\newcommand{\OUT}[1]{}
\usepackage[utf8]{inputenc}

\title{Conversational Assistants to support Heart Failure Patients: \\
comparing a Neurosymbolic Architecture with ChatGPT}

\author{
Anuja Tayal\textsuperscript{1} \quad
Devika Salunke \textsuperscript{2}\quad
Barbara Di Eugenio\textsuperscript{1} \quad 
Paula G Allen-Meares \textsuperscript{3} \\
\textbf{
Eulalia P Abril \textsuperscript{4} \quad
Olga Garcia-Bedoya\textsuperscript{3} \quad
Carolyn A Dickens\textsuperscript{3} \quad
Andrew D. Boyd\textsuperscript{2}} \\
\textsuperscript{1}Department of Computer Science 
\textsuperscript{2}Department of Biomedical and Health Information Sciences \\
\textsuperscript{3}Department of Medicine
\textsuperscript{4}Department of Communications \\
\texttt{\{atayal4,dsalun2,bdieugen,pameares,eulalia,ogarciab,cdickens,boyda\}@uic.edu} \\
}

\begin{document}
\maketitle
\begin{abstract}
Conversational assistants are becoming more and more popular, including in healthcare, partly because of the availability and capabilities of Large Language Models. There is a need for controlled, probing evaluations with real stakeholders which can highlight advantages and disadvantages of more traditional architectures and those based on generative AI. We present a within-group user study to compare two versions of a conversational assistant that allows heart failure patients to ask about  salt content in food. One version of the system was developed in-house with a neurosymbolic architecture, and one is based on ChatGPT. The evaluation shows that the in-house system is more accurate, completes more tasks and is less verbose than the one based on ChatGPT; on the other hand, the one based on ChatGPT makes fewer speech errors and requires fewer clarifications to complete the task. Patients show no preference for one over the other. 
\end{abstract}
\OUT{the systems, where patients were asked to query about salt content in 5 food items they consume every day. This user study was done not only to evaluate the 2 systems but also to gauge the eating habits of the AA population, how they describe foods, and how much they are aware of the salt amounts in different foods. For the user study, one of the systems being used in ChatGPT-4 would be the first system to test with real patients in a real setting. }

\section{Introduction}

Conversational assistants in the healthcare domain  are as old as  Natural Language Processing (NLP), since in 1966 ELIZA was already playing the role of a psychiatrist \cite{eliza}. They have proliferated in more recent years with the availability of datasets and machine learning approaches, even before the rise of  Large Language Models (LLMs). The comprehensive survey \OUT{of NLP literature conducted} in \cite{healthcare-dialog-survey} provides an in-depth analysis of these diverse healthcare-oriented dialogue systems, examining them from a computational perspective and highlighting their varied end-users. 

\OUT{As chatbots developed, the need for robust evaluation frameworks became apparent. Paradise \cite{paradise} was one of the early papers that provided proper guidelines to evaluate a dialog system that assess system performance by linking user satisfaction to task success and interaction costs. }

Traditional task-oriented dialog systems are typically assessed using metrics such as Slot Filling F1-Score and Intent Recognition Accuracy for Natural Language Understanding (NLU), as well as Slot Accuracy and Joint Goal Accuracy for Dialog State Tracking (DST) \cite{multiwoz}. However, task performance is only one aspect of dialogue system evaluation, as already demonstrated by Paradise \cite{paradise}, a framework that links user satisfaction to task success and interaction costs.  

Evaluation of dialogue systems based on LLMs raises additional concerns: LLMs do not operate within rigid task boundaries, making it difficult to apply standard task-oriented evaluation metrics. Furthermore, LLMs often lack transparency regarding their data sources and may fail to reliably follow user prompts, raising significant concerns in high-stakes domains where accuracy and accountability are critical \cite{llm-medical-evaluation}. These limitations are especially critical in practical, real-world applications when facilitating medical conversations. In such cases, defining and assessing what constitutes "good" performance becomes far more complex and nuanced \cite{llm-medical-weakness}.


 As a result, human evaluation remains the gold standard for assessing medical dialog systems \cite{llm-medical-evaluation}. Human evaluators can provide insights into subjective qualities such as coherence, informativeness, and user satisfaction—factors that are difficult to capture with automated metrics alone. 
 We present a user study in which we compare one traditional task-oriented dialog system (ToDS) in healthcare with one based on LLMs.

The study focuses on African American patients with heart failure, aiming to assist them in managing their salt intake by providing information about the salt content in various foods. 
Heart failure patients must meticulously monitor and reduce their salt intake:  African American individuals  are more prone to heart failure \cite{hfaa}, have a higher sensitivity to salt, and face challenges like lack of access to healthy foods
\OUT{food deserts and higher consumption of junk foods}. \OUT{This necessitates a specialized dietary management approach to help them monitor and reduce their salt intake effectively. } Furthermore, in \cite{itika-heart}, the authors show that African American patients with heart failure often \OUT{focus on discussions related to} discuss salt and food during heart failure educational sessions, indicating a significant interest and need for information in this area. By providing a tool that facilitates easy access to information about salt content in foods, we aim to empower patients to make healthier dietary choices, thereby addressing a critical aspect of managing heart failure.

The study compares two dialog systems—an in-house Neuro-Symbolic System (HFFood-NS) with a ChatGPT-based system (HFFood-GPT)—using a within-subject design. We conduct intrinsic (task performance)  and extrinsic analyses \cite{jones1995evaluating} using pre- and post-interaction surveys, to evaluate the 2 systems with African-American patients while hospitalized. 
By assessing how real patients, rather than typical study participants like Mechanical Turk workers or students, discuss food and prefer to receive information, we establish relevance and value of the intervention with real stakeholders.

\section{Related Work}
\textbf{ToDS.}
Task-oriented Dialog Systems (ToDS.) typically follow a pipeline approach with four main components: natural language understanding (NLU) \cite{nlu}, dialog state tracking (DST) \cite{dst}, dialog manager \cite{dialog-policy}, and natural language generation (NLG) \cite{nlg}. The NLU module interprets user input and represents the dialog state as slots (e.g., location, price range) to be filled during the conversation. DST monitors unfilled slots to inform the dialog manager, which decides the next action. This action is then passed to the NLG module to generate the system's response.

Early systems \cite{pompdp} used carefully designed action spaces to manage dialog states which were later replaced by neural networks \cite{sequicity, soloist}. While the pipeline approach integrates domain-specific knowledge and slot-filling methods, it often requires additional human labeling.

\OUT{BDE: the next paragraph is out of place / not corrected to rest. Cut it out}

\textbf{Dialog Systems in Healthcare.}  ToDS have seen a significant rise in the healthcare sector \cite{healthcare-dialog-survey}. These systems are developed for a wide array of diseases, including patient education \cite{panini, itika-heart}, heart failure \cite{dil,itika-heart}, mental disorders \cite{autism-ds}, public anxiety \cite{public-speaking-anxiety}, cancer \cite{cancer-ds} etc. Their applications extend to several areas, including disease diagnosis \cite{automatic-diagnosis-pipeline} and health coaching \cite{yue-health}, among others.

\textbf{Patient Centeredness.} \cite{itika-heart,hfchat} highlight the development of a dialogue agent tailored to the self-care needs of heart failure patients, leveraging insights from educational sessions. \cite{woz-chi-persona-health-counceling} investigates the use of the Wizard of Oz (WOZ) technique to create a persona-based health counseling dialogue dataset. Recent advancements have also incorporated Large Language Models (LLMs) to address patient inquiries \cite{llm-medical-evaluation}, with a strong emphasis on safety. To overcome the limitations in medical knowledge inherent to LLMs, \cite{chatdoctor} focuses on enhancing and fine-tuning the LLaMa model using a dataset of approximately 100,000 patient-doctor dialogues. \cite{user-study-medical-note} used a user-centered approach to iteratively improve their medical note generation model with user feedback conducted via semi-structured interviews. 
User studies are useful for collecting real-world 
evaluations in domains lacking specific use-case data \cite{paradise,riveiro2021}.





\section{Two dialogue system architectures}

\subsection{HFFood-NS} 
The first system, HFFood-NS, is a neuro-symbolic conversational system adapted from (Reference Withheld)\OUT{\cite{tayal-neuro}}. As there is no conversational dataset related to food salt content, we created a template-based conversational dataset.  We utilized the USFDC (U.S. Food Data Central) \cite{abc} dataset, which provided detailed food descriptions along with their nutrient values. We annotated the dataset based on food slots of \textit{food}, \textit{cook}, \textit{type}, \textit{foodweight}, and \textit{metric}. \OUT{and a food ontology built with FoodOn \cite{foodon}}Full details of creating the dataset can be found in (Reference Withheld). Initially, we trained an end-to-end dialog system, but for the user study, we only used its DST part. 
\subsubsection{End-to-End dialog system}
\label{e2e}
We used the end-to-end dialog system PPTOD (Plug-and-Play Task-Oriented Dialogue System) \cite{pptod} to train the dialog system. PPTOD, a T5-based model, excels in in-context learning by employing customized prompts for specific tasks.

Our experiments revealed that fine-tuning a transformer model to predict salt content alone is insufficient (see Table~\ref{table:full-model}). This is due to two key reasons. First, large pre-trained language models (PLMs) like GPT-3 and T5 \cite{few-shot,t5} occasionally make calculation errors, particularly as mathematical operations in equations grow more complex \cite{cot-reasoning}. Second, salt is a multi-valued slot, and its value changes over slight changes in food preparation method or quantity. 


To address these challenges, we integrated the PPTOD model with neuro-symbolic rules. These rules enable the retrieval of accurate salt values from a database and perform mathematical calculations for specific food weights, allowing the system to handle non-standard food quantities effectively. This integration significantly enhances system performance, achieving a 20\% improvement in joint goal accuracy across different dataset sizes (as shown in Table \ref{table:joint-accuracy}). These results demonstrate that combining pre-trained language models with neuro-symbolic rules 
achieves better accuracy.

\begin{figure}[t]
    \scriptsize
    \includegraphics[width=\linewidth]{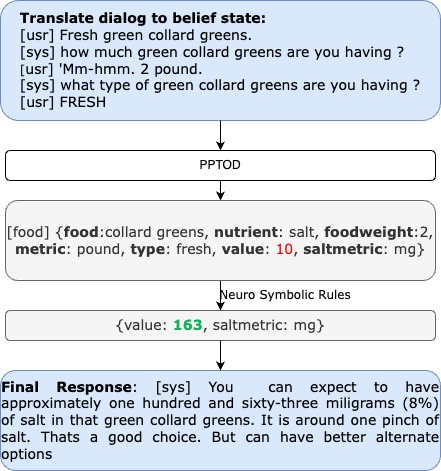}
    \caption{HFFood-NS with Interaction: As the value in DB was available for 100 gms, while the user asked about 2 pounds, it fetched the appropriate value and calculated the value for 2 pounds.}
    \label{fig:NSPPTOD}
\end{figure}


\subsection{Final Model- HFFood-NS}

For the user-study version of HFFood-NS, several measures were implemented to modify the system to deploy in a patient-centric application. First, only the Dialogue State Tracking (DST) module of NS-PPTOD was used for simplicity and reliability. Secondly, if a slot remains unfilled, the system attempts to query the user about it up to two times. Third, the system responses were template-based responses which were created to train the model (Section \ref{e2e}). Fourth, since the USFDC dataset contains detailed information and users often give less precise food descriptions, the system selects the first food item that meets all minimum required keywords to account for the multi-valued nature of food items when providing salt values.

Finally, we carefully design the final system response to effectively represent salt values. We represent salt values using two formal units—milligrams (mg) and percentage (\%)—commonly found on food labels, and we include an informal measure - pinches/dashes. We also compared the salt value to the recommended daily intake of 2000mg.  We structure the final response to be motivational by categorizing salt content into four ranges (<5\%, <20\%, <50\%, <100\%, and others) and crafting a motivational message. Figure \ref{fig:NSPPTOD} shows an example of a user-study interaction and the application of the neuro-symbolic rule.


\subsection{HFFood-GPT}

We used assistant GPT to create HFFood-GPT as it was easier to integrate with the UI. Since our goal was to educate patients without providing medical advice, we prompted it (the prompt is provided in Appendix \ref{app:gpt4-prompt}) to not give health advice or suggest consulting a professional for dietary guidance. We named the system Sodium Scout and prompted to analyze the salt contents in foods. It advises that foods exceeding 20\% of this intake are not recommended, while those below 5\% are favorable choices. We enabled the code interpreter and retrieval features to enable data access. This allowed the assistant to retrieve information from the USFDC dataset, which was also used to create the HFFood-NS dataset. We instructed it not to mention the dataset with patients or search the web for information. 

To compare HFFood-GPT with our in-house system, HFFood-NS, we prompted GPT-4 \cite{gpt-4} to function as similarly as possible to HFFood-NS. We prompted Sodium Scout to ask clarification questions about food type, cooking method, and portion size, limiting the questions to one question at a time. Additionally, to address GPT-4's tendency to produce lengthy responses, we prompted it to keep the responses under 40 words.



\section{User Study}
\textbf{SetUp}  We recruited 23 African American (AA) patients, aged between 18 and 89 years, who could speak English and had a history of heart failure. All participants were compensated for their time. The study was conducted while they were hospitalized, in their hospital bedroom.

A total of 20 participants (13 Males and 7 females, ages  18-89 - $\mu =$ 58.75, $\sigma = $ 14.32) completed the study which took around 6 months. Proper IRB guidelines were followed. If the participants met the inclusion criteria and were willing to participate in the study, they were provided with an informed-consent document and pre-survey questions.  Each participant interacted with the two versions of the dialog agent (DA) one after the other in a randomized manner to reduce recall bias. 

 To help the participants think of questions, we prompted the participants to think about foods they ate for breakfast, lunch, or dinner. Participants then asked questions about the same food items to both dialog systems. This was done to ensure that the participants did not run out of questions for the second interaction and to make it easier to compare the 2 DAs.
However, the interactions were not identical because each system asked different clarification questions, which led the conversations in different directions.

In this study, all interactions with the systems were conducted orally -  the dialog systems were speech-based. This included obtaining oral consent and administering both the pre-and post-survey questions verbally. The decision to use oral communication was made to accommodate participants who might be unwell, or be connected to medical equipment, ensuring a more accessible and comfortable experience.

Conducting a user study with hospitalized patients comes with unique challenges and considerations. For instance, Patient 15's partner, who was the primary caretaker responsible for managing food, was also present for the session and together asked questions. Patient 17, despite being visually impaired, had no difficulty participating in the study as the system was speech-based. One patient had a tracheostomy tube, and another was pregnant, highlighting the diversity of participants. Additionally, three patients initially faced some challenges in understanding the systems due to their pain but eventually became comfortable in using it.


\textbf{UI} 
The UI was minimalistic and designed for ease of use, featuring two radio buttons to select the system 
\OUT{(Lion for the HFChat Model and Shark for the GPT-4-based model) } and a button to indicate whether the system was recording or listening.  For the UI, we utilized Gradio \cite{gradio}, a framework provided by Hugging Face, as it offers a simple and effective solution. \OUT{ We kept the interface consistent across both systems, with the only difference being the underlying architecture.} 


The UI was not "always listening" and could only listen or speak one at a time. It was done to prioritize patients' health and accommodate the presence of nurses and doctors during interactions. We named the two systems Lion and Shark so that the participants could easily recall the names of the systems being used. The two names were chosen as they represent a well-known, powerful animal.

We utilized OpenAI technologies for both Text-to-Speech (TTS) and Speech-to-Text (STT) functionalities in the study. For Speech-to-Text, we employed Whisper-1 \cite{whisper}, and for Text-to-Speech, we used TTS-1. We recorded audio and collected transcripts for  error analysis. 

\OUT{to calculate the Word Error Rate (WER) later, ensuring a detailed evaluation of the system's performance. During testing, we found that Whisper-1 occasionally misinterpreted 4-digit numbers, which was resolved by converting them into their textual representation. Additionally, while Whisper automatically detects the speaker's language, it sometimes misidentifies the language as something other than English, even when the participants were speaking in English. To address this, we fixed the language setting to English, as the study was conducted entirely in English.}


  

\begin{figure}[ht!]
  \centering
  \subfigure[Health Literacy Vs Preference]{%
    \includegraphics[width=0.4\textwidth]{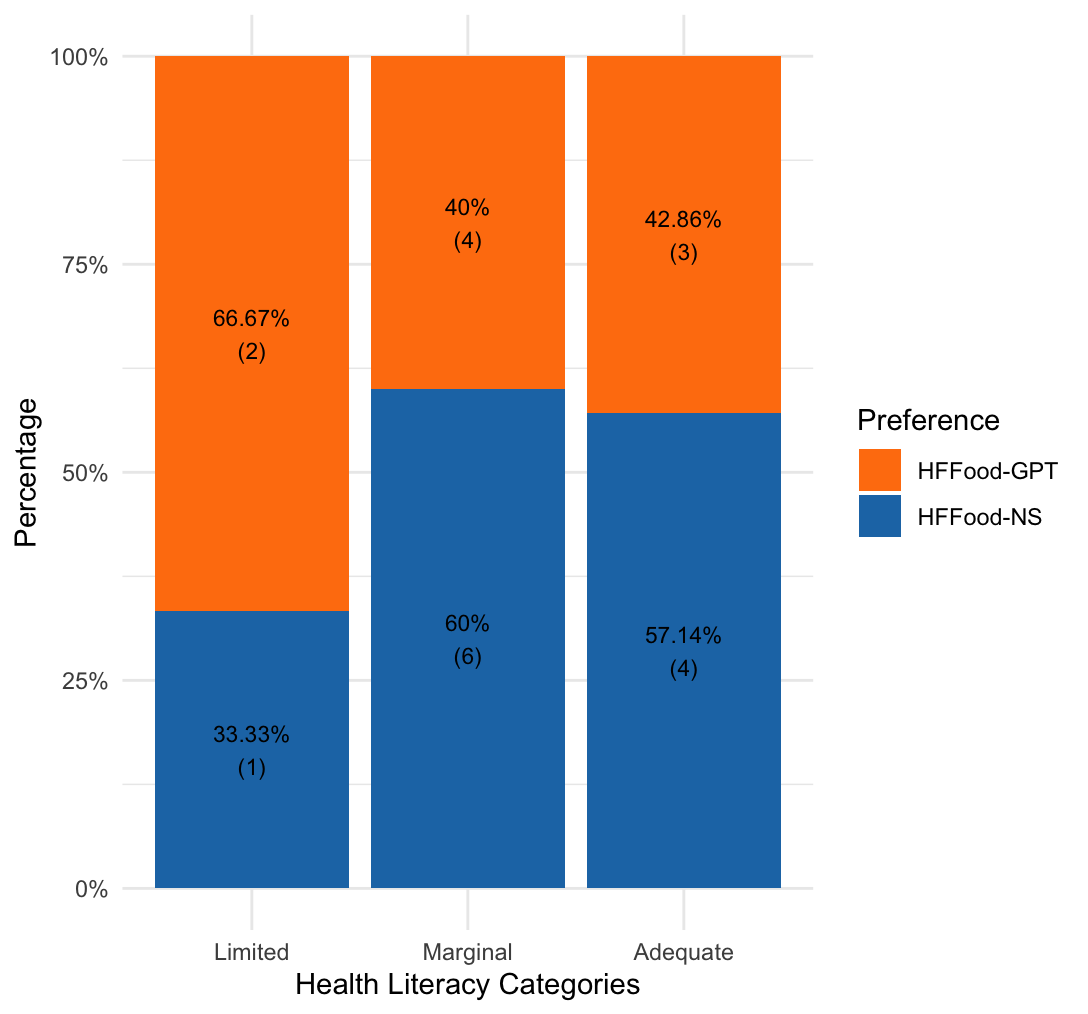}
  }
  \subfigure[Digital Health Literacy Vs Preference]{%
    \includegraphics[width=0.4\textwidth]{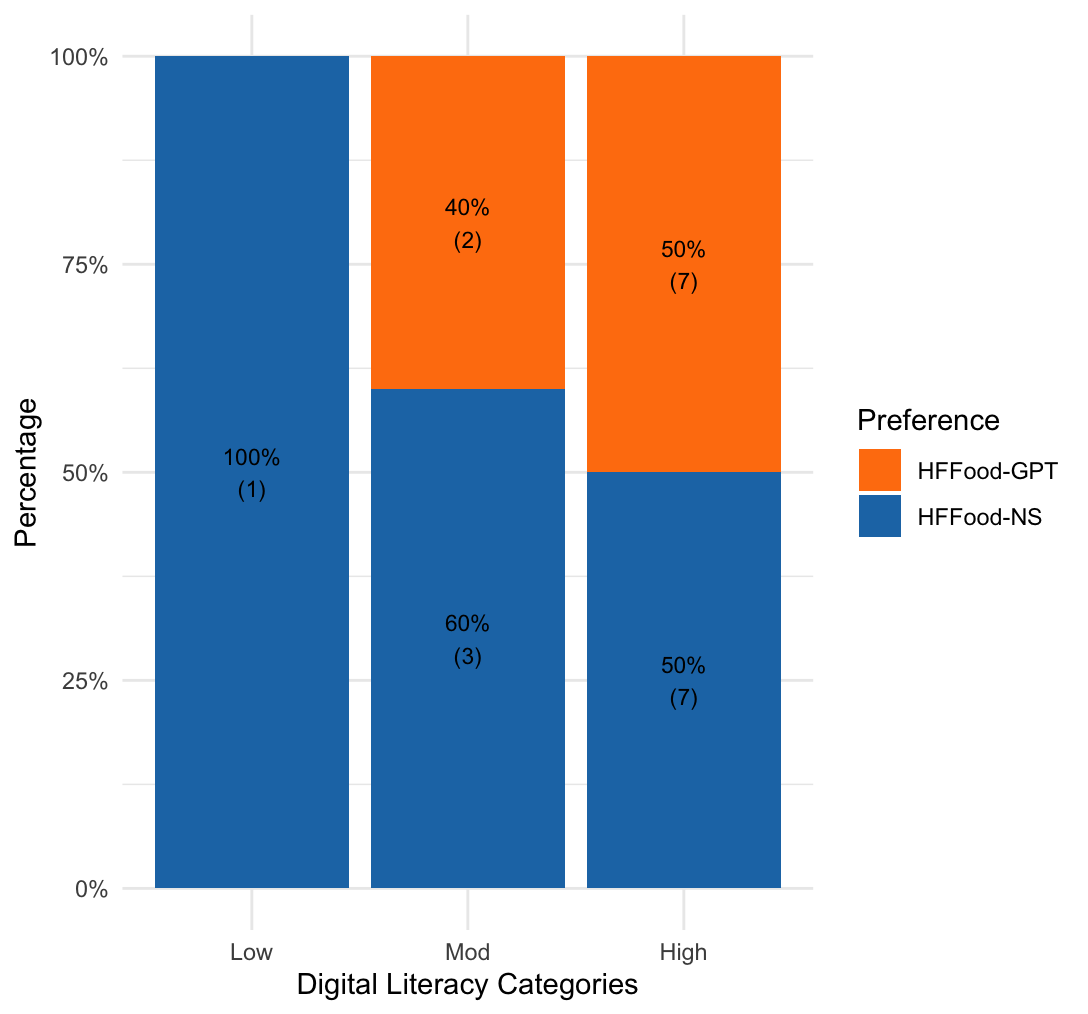}
  }
  \caption{Comparison of user preferences categorized by health literacy and digital health literacy levels, with numbers in brackets indicating the number of users.}
  \label{fig:pre_survey_questionnaire}
\end{figure}

\subsection{Survey Questions}
\subsubsection{PreSurvey Questionnaire}
We assessed participants' health literacy and digital literacy using self-reported measures to better understand their abilities to access and process health information. The PreSurvey Questionnaire can be assessed in Appendix \ref{app:pre-survey}.
\OUT{These assessments helped us understand their interactions with the systems based on their literacy levels.} 

\textbf{Health Literacy} was measured using the \textbf{BRIEF Health Literacy Screening Tool} \cite{brief}.  This tool consists of 6 questions that evaluate individuals' ability to "obtain, process, and understand basic health information and services needed to make appropriate decisions about their health."

\textbf{Digital Literacy} was measured using the \textbf{eHealth Literacy Scale (eHEALS)} \cite{eHeals}. This tool consists of 8 questions that assess participants' perceived ability to find, evaluate, and apply electronic health information to address health-related issues. 


\subsubsection{PostSurvey Questionnaire}
After users interacted with both systems, we asked them to evaluate their experience through a questionnaire. The Post-Survey Questionnaire is available in Appendix \ref{app:post-survey}.

We asked users about the following aspects of the two systems: how easy it was to understand the answers, the usefulness of the questions, and how easy the questions were to understand. We also asked which system they preferred, the most challenging aspect of using them, their likelihood of using the system in the future, whether the system was helpful, and if they had any additional feedback.


\section{Evaluation}

\subsection{Intrinsic Evaluation}
We used the USFDC dataset as a baseline to evaluate the accuracy of the two systems. We treated HFFood-GPT as a task-oriented dialogue system by categorizing the provided values into defined slots (Food, Cook, Type, FoodWeight, and metric) similar to HFFood-NS.

\begin{table}[ht]
\scriptsize
    \begin{tabular}{c|c|c}
    \hline
          & \textbf{HFFood-NS} & \textbf{HFFood-GPT} \\
          \hline
         \textbf{Avg No of turns} & 3.6 & 3 \\
         \textbf{Avg Processing Time} & 6.7 & 11.4 \\
         \textbf{Avg Words of the System} & 14.5 & 54.5 \\
         \textbf{Avg Retries} & 2 & 1.7 \\
         \textbf{Avg WER} & .483 & .41 \\
         \textbf{Task Completion} & 84\% & 62\% \\
         \textbf{Accuracy} & 37\% & 24\% \\
    \hline
    \end{tabular}
    \caption{Intrinsic Evaluation: Comparing 2 systems}
    \label{table:Analysis}
\end{table}

Table~\ref{table:Analysis} summarizes the analysis of task completion and accuracy for both systems. \textbf{Task completion} is defined as the system providing a value, whether correct or incorrect, while \textbf{accuracy} refers to the system providing the correct salt value. Speech errors were excluded from the accuracy calculation. For task completion, we considered cases where the model provided a salt value (either as a specific number or a range) rather than categorical descriptors as observed in HFFood-GPT.

Evaluating HFFood-GPT proved particularly challenging due to its black-box nature. To maintain consistency, we evaluated it similarly to HFFood-NS by comparing the provided salt value to the first food item in the USFDC database that satisfied all minimum slot values. However, it remains unknown which data HFFood-GPT actually accesses or the process it follows to calculate its final answers.

Although HFFood-GPT demonstrated lower task completion accuracy compared to HFFood-NS, it achieved higher slot accuracy. Table \ref{table:hfchat-slot-analysis} and Table \ref{table:gpt-slot-analysis} present the slot accuracy analysis for HFFood-NS and HFFood-GPT, respectively. We classified the incorrect slots into Speech Errors and Partial Speech Errors (P-SE), which likely occurred due to the noisy hospital environment, patients' accents, or the overall conditions in a hospital setting.

Although we used the same Text-to-Speech and Speech-to-Text systems for both the systems, HFFood-NS recorded a higher word error rate (WER) \cite{wer} than HFFood-GPT (Table \ref{table:Analysis}). HFFood-GPT's incorrect slot errors resulted exclusively from Speech Errors and Partial Speech Errors.


\begin{table}[ht]
\scriptsize
    \begin{tabular}
    {c|c|c|c|c}
    \hline
          & \textbf{Correct} & \textbf{Incorrect} & \textbf{Speech Error} & \textbf{P-SE} \\
          \hline
         \textbf{Food} & 86 & 6 & 4 & 1 \\
         \textbf{Cook} & 39 & 21 &11 & 2 \\
         \textbf{Type} & 44 & 31 &17 & 2\\
         \textbf{Foodweight} & 29 & 56 &18& 0\\
         \textbf{Metric} & 27 & 60 &20& 0\\
    \hline
    \end{tabular}
    \caption{HFFood-NS Slot Accuracy Analysis: incorrect includes Speech Error and Partial Speech Error (P-SE)}
    \label{table:hfchat-slot-analysis}
\end{table}  


\begin{table}[ht]
\scriptsize
    \begin{tabular}
    {c |c | c | c|c}
    \hline
          & \textbf{Correct} & \textbf{Incorrect } & \textbf{Speech Error } & \textbf{P-SE}\\
          \hline
         \textbf{Food} & 94 & 7 &3 & 4  \\
         \textbf{Cook} & 46 & 5 & 5 &0  \\
         \textbf{Type} & 58 & 15 & 9 &6  \\
         \textbf{FoodWeight} & 59 & 5 &5 &0 \\
         \textbf{Metric} & 59 & 5 & 5 &0 \\
    \hline
    \end{tabular}
    \caption{HFFood-GPT Slot Accuracy Analysis: incorrect includes Speech Error and Partial Speech Error (P-SE)}
    \label{table:gpt-slot-analysis}
\end{table}

\subsection{Extrinsic Evaluation: User Perceptions}

\begin{figure}[t]
    \scriptsize
    \includegraphics[width=\linewidth]{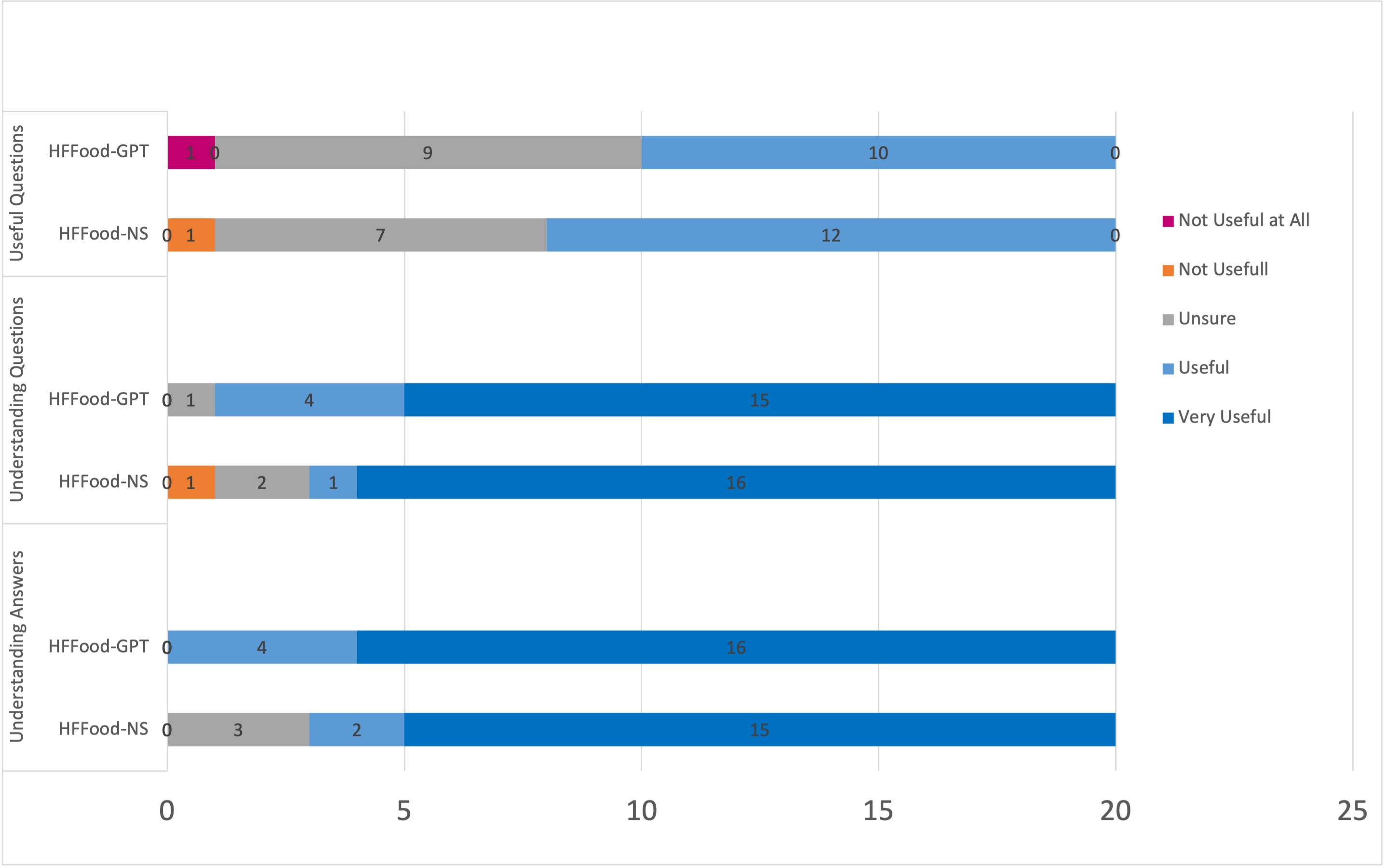}
    \caption{Post-survey: extrinsic evaluation of the two systems on the usefulness and understanding of answers and questions.}
    \label{fig:post-survey}
\end{figure}

Figure \ref{fig:post-survey} shows how participants rated the two systems in the post-survey questionnaire, focusing on Understanding Answers, Understanding Questions, and Useful Questions. We can see that more than 50\% of the people find all the aspects useful.

 Out of 20 patients, 11 preferred HFFood-NS, while 9 favored HFFood-GPT.  Since this is a within-group study, there is a potential for recall bias; however, we did not observe any evidence of it. In 10 cases where HFFood-NS was tried first, 7 users favored it, while in 10 cases where HFFood-GPT was tried first, 6 users favored it. But, the preference for the first system was not statistically significant, as indicated by a Pearson correlation (r = 0.3, p = 0.19).

\OUT{Among the 10 cases where both systems were tried first, 7 cases favored the first system for HFFood-NS, and 6 cases favored the first system for HFFood-GPT} 

A comparison of preferences based on health and digital health literacy, as assessed through the pre-survey questionnaire, is presented in Figure~\ref{fig:pre_survey_questionnaire}. The numbers are too small to draw any definite conclusions.

When we asked the users about the reasons for their preferences, those who favored HFFood-NS highlighted its precise and to-the-point answers, faster flow, \OUT{additional information} and concise responses. On the other hand, users who preferred HFFood-GPT appreciated the ease of understanding its questions, clear and detailed explanations, and better-formulated questions.

While some users found no issues with either system and felt they understood them well, others reported challenges. These included the systems being repetitive, difficulty understanding the questions, and uncertainty about how to phrase their own questions.

\paragraph{Would patients use the DA in their daily life?}

\begin{figure}[t]
    \scriptsize
    \includegraphics[width=\linewidth]{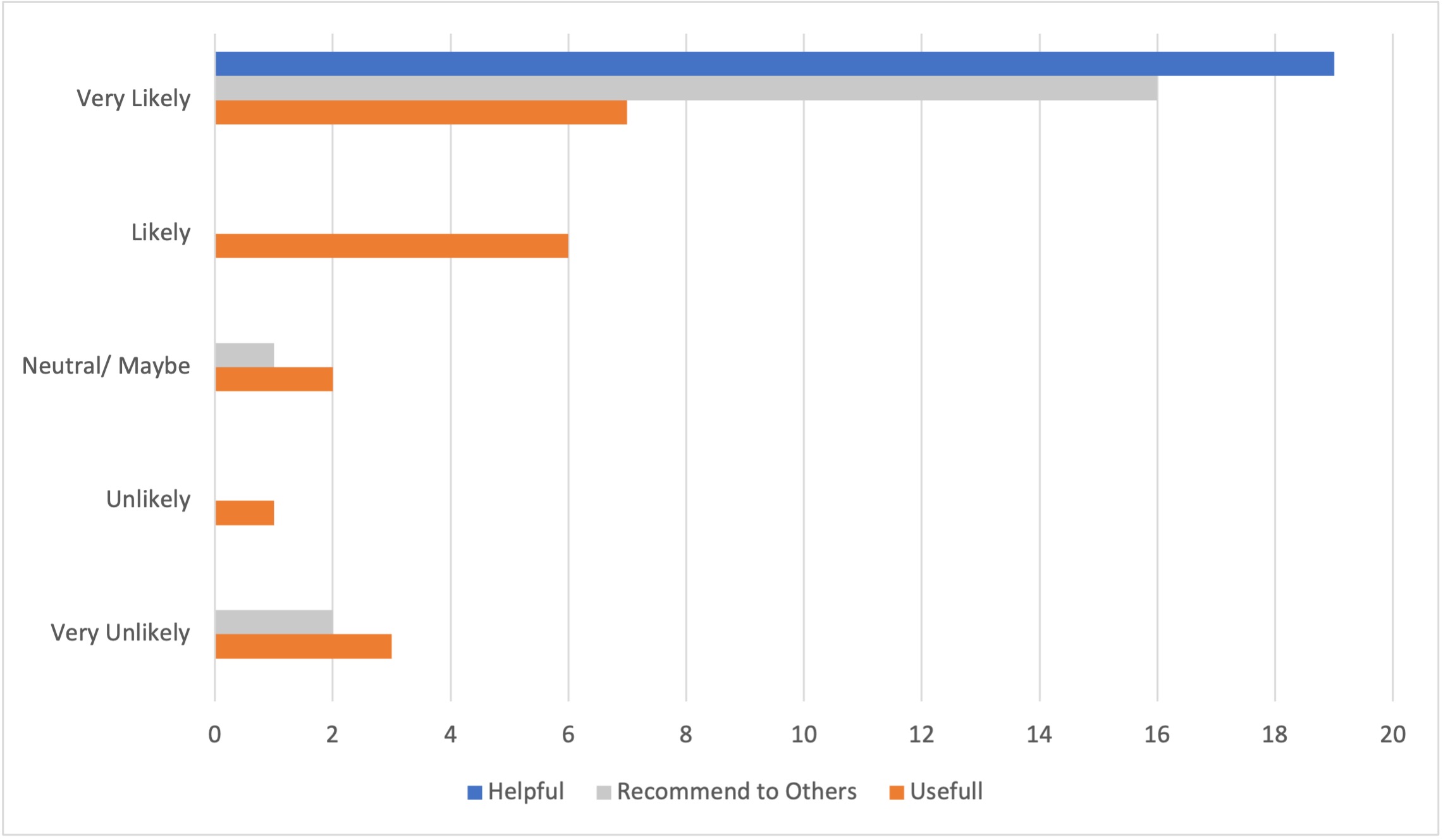}
    \caption{Post Survey Analysis: How users perceived the conversational system}
    \label{fig:user-perceive-ds}
\end{figure}

\OUT{We had to stop the post-survey questions in between for patient p5 as there was a procedure scheduled. }
Figure \ref{fig:user-perceive-ds} highlights that all users found the system helpful, with most indicating they would recommend it to others. The majority found both systems useful. However, data is only available for 19 patients, as the post-survey questions for patient P5 were interrupted due to a scheduled procedure.

\OUT{When we asked users how often they would use such a system, they responded with answers such as every day, once a week, quite often, as needed, or whenever cooking.}

\paragraph{Qualitative measures of salt content.}
In the post-survey, 55\% of users (11 participants) preferred receiving information about salt content in informal terms, such as pinches or dashes. Only 20\% (4 participants) preferred formal units like milligrams or grams, while the rest chose percentages or daily values. In the pre-survey, participants reported measuring salt informally by sprinkling, using pinches, or judging by eye. These informal methods differ from the formal values, such as milligrams or percentage of daily value, that appear on food labels.

\section{Discussion: comparison between the two systems}

Table \ref{table:comparison} summarises the 2 systems, highlighting the pros and cons by comparing performance, Design and Usability, reliability, and flexibility.

\begin{table}[ht]
\scriptsize
    \begin{tabular}{c|c|c}
    \hline
          & \textbf{HFFood-NS} & \textbf{HFFood-GPT} \\
          \hline
         \textbf{Task completion} & \cmark & \xmark \\
         \textbf{Accuracy} & \cmark & \xmark \\
         \textbf{Slot Accuracy} & \xmark & \cmark \\
          \textbf{Less Speech Error} & \xmark & \cmark \\
          \textbf{Less Processing Time} & \cmark & \xmark \\ \hline
          
          \textbf{More Constrained} & \cmark & \xmark \\
         \textbf{Error Analysis} & \cmark & \xmark \\
          \textbf{Reliable} & \cmark & \xmark \\
           \textbf{Predictable} & \cmark & \xmark \\
           \textbf{Handling Complex query} & \xmark & \cmark \\ \hline

         \textbf{Gave Options} & \xmark & \cmark \\
          \textbf{Fluent} & \xmark & \cmark \\     
          \textbf{Concise} & \cmark & \xmark \\
         \textbf{Create easily with less time} & \xmark & \cmark \\

    \hline
    \end{tabular}
    \caption{Pros and Cons of HFFood-NS and HFFood-GPT comparing on performance, Design and Usability, reliability and flexibility.}
    \label{table:comparison}
\end{table} 

\subsection{HFFood-NS}
One significant advantage of employing a neuro-symbolic system combined with a traditional ToDS, rather than directly prompting a LLM, lies in the ability to conduct more thorough and precise error analysis. This capability is crucial, particularly in patient-centric systems. By utilizing this approach, we were able to accurately pinpoint the areas where our system fell short.

Table \ref{table:error-analysis} provides a detailed breakdown of the error analysis performed on HFFood-NS. This analysis includes multiple overlapping categories, which highlight the nuanced nature of errors encountered. By systematically addressing these issues, we can iteratively develop a more robust and reliable dialogue system.

\begin{table}[ht]
\scriptsize
\centering
    \begin{tabular}{c|c}
    \hline
          & \textbf{HFFood-NS}  \\
          \hline
         \textbf{Missed Slot} & 27  \\
         \textbf{Wrong Food Identified} & 9  \\
         \textbf{System Error} & 8  \\
         \textbf{Internet} & 6  \\
         \textbf{No Data Fetched} & 5 \\
         \textbf{Wrong Math} & 5  \\
         \textbf{Food Not in USFDC} & 4  \\
         \textbf{Complex Query} & 3  \\
         \textbf{Missed Slot not in USFDC} & 2  \\
         \textbf{Wrong Food Fetched} & 1 \\
    \hline
    \end{tabular}
    \caption{Through and precise error analysis HFFood-NS}
    \label{table:error-analysis}
\end{table}

Moreover, neuro-symbolic rules gave us the ability to add fail-safe to the system where when the model was not able to infer the slot for food weight and metric, it would assume 100g, thereby increasing the task completion rate.  

Moreover, having greater control over the system provided significant advantages. Firstly, it ensured that the system remained aligned with its primary goal, which HFFood-GPT struggled to maintain consistently. For example, in one instance, HFFood-GPT asked a clarification question about the color of the bell pepper being used—a detail that is considered irrelevant when determining the salt amount. This level of control helped minimize distractions and kept the system focused on delivering accurate and goal-oriented responses.

Secondly, ensuring consistency in responses by providing values in milligrams (mg) made the system more user-friendly and reliable. In contrast, HFFood-GPT showed some inconsistency in its responses, with 10\% of values given in grams (g), 10\% as categorical descriptors (e.g., low, high, significant), 8\% as ranges instead of precise values, and 2\% in teaspoons (tsp) and rest in mg. This inconsistency could pose challenges for patients, 
who may overlook that 1 gram is equivalent to 1000 milligrams due to varying levels of numerical literacy \cite{foodlabel}. Maintaining consistency in responses is a crucial step to improving accessibility and ensuring users can easily interpret the information provided.

\subsection{HFFood-GPT}

HFFood-GPT handled diverse user descriptions effectively, interpreting non-standard inputs and maintaining smooth conversations. Patients described their food portions in various unconventional ways. Many reported portion sizes in terms of servings, such as "for 6 people" or "serving size for 1," instead of using standard units like ounces, grams, or pounds. Two patients used unique measurements, such as "spoonfuls" and "forkfuls," while two of them relied on size comparisons, like describing a hotdog as "the size of a cellphone" or quantifying food weight as "3 inches of bacon". HFFood-GPT demonstrated flexibility in handling these inputs, whereas HFFood-NS struggled with slot identification, as reflected in Table \ref{table:hfchat-slot-analysis}, where most errors occurred in food weight and metric slots.


When HFFood-GPT was unable to determine the salt amount in a food item directly, it attempted to gather the recipe and estimate the salt content based on the salt levels of its ingredients. While this approach was sometimes helpful, there were instances where the system asked overly specific questions, such as the exact weight or brand of each ingredient, which could be burdensome for users.

In some cases, the brand names were so widely recognized that the food was identified by the brand itself, such as "Chick-fil-A" or "Hurricane Whopper." In these situations, HFFood-GPT successfully deduced the food item without requiring additional descriptive details, showcasing its ability to handle commonly branded food references effectively.

However, due to the unpredictable nature of HFFood-GPT, the system sometimes specifically asked for food weight in grams or ounces. For example, in the conversation shown in Table \ref{table:conv2}, when asked about weight in grams, the user was not aware of it. It was good that HFFood-GPT tried to educate about it. Users may not always know these exact measurements and are more likely to provide quantities in terms they are comfortable with.


\begin{table}[ht]
\scriptsize
    \begin{tabular}{p{.05 \textwidth}|p{.37 \textwidth}}
    \hline
          & \textbf{Turn} \\
          \hline
         \textbf{System} & Thank you for specifying the preparation method. Now, could you please tell me the portion size of the grilled turkey sausage in grams? \\
         \textbf{User} & 5 milligrams. \\
         \textbf{System} & It seems there might be a misunderstanding with the portion size. Typically, food portion sizes are measured in grams (g) and would be much larger than 5 milligrams (mg). Could you please provide the portion size again in grams? For reference, a common serving size for sausage might be around 50 to 150 grams. \\
         \textbf{User} & 100 milligrams \\
         \textbf{System} & It looks like we're still not quite accurate on the portion size. When we're talking about turkey sausage, the realistic portion size is usually around 100 grams (g), not milligrams (mg). One gram is equal to 1000 milligrams, so a portion size of turkey sausage might be 100 grams (which is 100,000 milligrams), for instance. Could you clarify the portion size in grams for the turkey sausage? \\
         
    \hline
    \end{tabular}
    \caption{Excerpt of a HFFood-GPT Conversation}
    \label{table:conv2}
\end{table}

There were 8\% instances where HFFood-GPT attempted to assume the slots instead of asking a clarification question. While this approach sometimes aligned with user intent, the goal should be to avoid making assumptions and instead provide users with the choice. For example, in one case, the user mentioned "chicken breast," and despite a potential speech error, the system correctly identified it because chicken breast is the most common way to consume chicken. However, if the user had been asking about "chicken tenders," this assumption would have resulted in an incorrect response.


When building HFFood-GPT, we set specific rules to constrain its behavior. We limited responses to 40 words, but as shown in Table \ref{table:Analysis}, the system averaged 54.5 words per response. We also instructed it to ask only one question per turn, but it violated this rule in about 7\% of its turns by asking multiple questions at once.

We had prompted HFFood-GPT to only refer to the database provided and not refer it to the users. However, in 38.5\% of system utterances, HFFood-GPT referenced the DB and said that the food item was not in the referenced DB and asked for more clarification. This could be one of the many reasons users did not prefer HFFood-GPT, as it could be off-putting. When we asked users how they would like to improve the system, their primary suggestions were to expand the knowledge base, include more ingredients, and add a broader range of food items to make the system more comprehensive and user-friendly.

GPT performed well by offering users options, such as saying, \textit{"That sounds delicious! Could you tell me if you're using fresh or dried spaghetti?"} which happened in around 30\% of system turns. This showed that HFFood-GPT had knowledge about different foods. However, in some cases, even after the option was chosen, HFFood-GPT responded that it lacked the information in the provided database. This behavior frustrated users because GPT relied on its own knowledge instead of the dataset, compromising the system's consistency and reliability.

\section{Conclusion and Future Work}
We conducted a user study with 20 African-American Heart Failure hospitalized patients. Using a within-subject design, we compared two dialogue systems: an in-house NeuroSymbolic System (HFFood-NS) and a ChatGPT-based system (HFFood-GPT).
The evaluation utilized both intrinsic and extrinsic measures, and while neither system emerged as a clear winner, the study highlighted key differences between the two.

The evaluation revealed that HFFood-NS is more accurate, completes more tasks, and provides concise responses compared to HFFood-GPT. On the other hand, HFFood-GPT makes fewer speech errors, requires fewer clarifications to complete tasks, and handles complex queries more effectively.

The widespread use of large language models (LLMs), like ChatGPT, often lacks scrutiny, raising concerns in healthcare settings. Greater control is needed, as relying solely on prompting is not enough. Neuro-symbolic methods, which offer greater transparency, reliability, and explainability, should be further explored and integrated into future systems.

Moving forward, we aim to develop hybrid conversational systems that combine the strengths of both systems.

\section{Limitations and Ethics Statement} 
While we believe user studies/ human evaluation are the best methods to assess medical dialogue systems, they have limitations. Our study recruited only 20 patients, which is a relatively small sample size, and it is difficult to get significant results.

To build the two systems, we relied on the USDA Food Composition Database (USFDC). While this database is a standard reference, it is neither fully exhaustive nor completely accurate. For instance, some foods, such as items from local restaurant chains or specific snack brands, are absent from the USFDC database. Additionally, in conversations about fried foods, users often referred to air fryer cooking methods, which were not accounted for in the database. 







\bibliography{sample}
\appendix

\section{NS-PPTOD Evaluation Results}

\begin{table}[t]
\scriptsize
    \begin{tabular}{|p{.1\textwidth} p{.07\textwidth} p{.07\textwidth} | p{.13\textwidth}|}
    \hline
           \textbf{Train Size} &  \textbf{Epochs} & \multicolumn{2}{c|} {\textbf{Joint Accuracy}} \\
           & & \textbf{PPTOD} & \textbf{NS-PPTOD} \\
         \hline
         100&6&55.56&73.08 \\
         300&4&51.92&72.8 \\
         500&6&58.75&83.2 \\
         1000&6&58.53 &85.2 \\
    \hline
    \end{tabular}
    \caption{Increase in Joint Accuracy when using NS-PPTOD compared to PPTOD across different training sizes}
    \label{table:joint-accuracy}
\end{table} 

\begin{table}[t]
\scriptsize
    \begin{tabular}{{p{.08\textwidth} p{.05\textwidth} p{.05\textwidth} p{.05\textwidth} p{.05\textwidth} p{.05\textwidth}}}
    \hline
           &  \textbf{Train Size} &  \textbf{Epochs} & \textbf{Inform} & \textbf{Success} & \textbf{BLEU} \\
         \hline

         PPTOD &100&8&71.43&0&24.99 \\
         NS-PPTOD &100&-&88.90&77.80&22.50\\
         \hline
         PPTOD &300&7&75.00&5.00&34.30 \\
         NS-PPTOD &300&-&81.50&63.00&26.90\\
         \hline
         PPTOD &500&9&82.86&2.86&29.81 \\
         NS-PPTOD &500&-&74.50&58.10&28.90\\
         \hline
         PPTOD &1000&7&93.50&2.70&29.00 \\
         NS-PPTOD &1000&-&85.90&71.70&30.00\\
         
    \hline
    \end{tabular}
    \caption{Increase in performance when using NS-PPTOD compared to PPTOD.}
    \label{table:full-model}
\end{table}  

\section{HFFood-GPT prompt }
\label{app:gpt4-prompt}

To develop HFFood-GPT, we used zero-shot prompting on GPT-4.

\textit{Sodium Scout greets users warmly and helps analyze the salt content in various foods. It methodically asks users about the\textbf{ food type, cooking method, and portion size, one question at a time, to accurately determine the salt content.} Using values from a provided JSON file, Sodium Scout calculates the estimate salt content and compares it to the recommended daily intake of 2000mg. It advises that foods exceeding 20\% of this intake are not recommended, while those below 5\% are favorable choices. \textbf{Sodium Scout refrains from giving health advice and suggesting from consulting a professional for dietary guidance.} The interface is friendly and straightforward. It focuses on informing users about salt levels in their meals with clarity, ensuring to ask only one question per turn. Answers are \textbf{kept under 40 words}, and it only searches the data provided in the JSON file. \textbf{Users do not know about the data file, so don't discuss it. Only focus on information related to food and their salt amount. Do not look for information on the web.}}


\section{Pre-Survey Questions}
\label{app:pre-survey}
\begin{itemize}
\item eHealth Literacy Questions
 \begin{enumerate}
     \item How useful do you feel the Internet is in helping you in making decisions about your health?
    \item How important is it for you to be able to access health resources on the Internet?
\item I know what health resources are available on the Internet
\item I know where to find helpful health resources on the Internet
\item I know how to find helpful health resources on the Internet
\item I know how to use the Internet to answer my questions about health
\item I know how to use the health information I find on the Internet to help me
\item I have the skills I need to evaluate the health resources I find on the Internet
\item I can tell high quality health resources from low quality health resources on the Internet
\item I feel confident in using information from the Internet to make health decisions
\end{enumerate}
\item Brief Health Literacy Screening Questions
\begin{enumerate}
    \item How often do you have someone help you read hospital materials?
    \item How often do you have problems learning about your medical condition because of difficulty understanding written information?
    \item How often do you have a problem understanding what is told to you about your medical condition?
\item How confident are you filling out medical forms by yourself? 
 \end{enumerate}    
    \item Do you pay attention to salt in your food [Yes, No, Maybe] How do you measure it?

\end{itemize}
\section{Post-Survey Questions}
\label{app:post-survey}
\begin{itemize}
    \item 1.	How easy was it to understand the answers you received from [insert system name: Lion/ Shark]? Please rate from 1 to 5, where 5 is very easy to understand and 1 is very difficult to understand.
\item	Did you think [insert system name: Lion/Shark] asked useful questions? (a.	Mostly Yes, b.	Yes, c.	Mostly No,d.	No)
\item	How easy was it to understand questions from [insert system name Lion/Shark]? Please rate from 1 to 5, where 5 is very easy to understand and 1 is very difficult to understand.

\item	How would you like to receive information about the salt content in your food. Would you prefer to see it as milligrams, as a percentage of your daily value, or in some other way? Please share your preference.

\item	Which system, Lion or Shark, do you prefer?

\item	In your own words, what was the hardest aspect of using the two systems?
\item	How likely are you to use such a system? (Rate 1 to 5 for each system: 1-Very unlikely to use, 2: Unlikely to use, 3: Neutral, 4: Likely to use, 5: Very likely to use) a.	Why?
b.	If the participant's response is yes, 
i.	How often would you use it? 
ii.	Would you recommend it to others? (Yes/No)
\item	Was the system helpful? a.	How would you improve the system? (If there is a preferred one), b.	How would you improve the systems? (If there is **not** a preferred one)
\item	Do you have any additional feedback or comments?

\end{itemize}

\end{document}